\documentclass{article}
\usepackage{arxiv/arxiv}
\usepackage{multicol}

\usepackage[utf8]{inputenc} 
\usepackage[T1]{fontenc}    
 \usepackage[colorlinks=true, citecolor=blue, urlcolor=blue, linkcolor=black]{hyperref} 
\usepackage{url}            
\usepackage{booktabs}       
\usepackage{amsfonts}       
\usepackage{nicefrac}       
\usepackage{microtype}      
\usepackage{lipsum}
\usepackage[english]{babel}
\usepackage{caption}
\usepackage{graphicx}
\usepackage{mdframed}
\usepackage{float}
\usepackage{hyperref}
\usepackage{amsmath,multirow,censor,amssymb, mathrsfs, dsfont, mathtools}
\usepackage[ruled,vlined]{algorithm2e}

\DeclareMathOperator*{\argmin}{argmin}

\usepackage{empheq}
\newcommand*\textfbox[2][Title]{%
  \begin{tabular}[b]{@{}c@{}}#1\\\fbox{#2}\end{tabular}}

\makeatletter

\newenvironment{figurehere}
  {\medskip\def\@captype{figure}}
  {\medskip}
\makeatother

\newcommand\fnurl[1]{%
\footnote{\url{#1}}%
}
\usepackage{tikz}

\usepackage[sort,authoryear,round]{natbib}
\title{\Large \textbf{An Introduction to Variational Inference}}
\author{\large Ankush Ganguly\thanks{Email: agang@sertiscorp.com} \and 
\large Samuel W. F. Earp \\ 
\AND {\large Sertis Vision Lab}\thanks{597/5 Sukhumvit Road, Watthana, Bangkok, 10110, Thailand}~
}

\date{}
\begin{document}

\maketitle

\begin{abstract}
Approximating complex probability densities is a core problem in modern statistics. 
In this paper, we introduce the concept of Variational Inference (VI), a popular method in machine learning that uses optimization techniques to estimate complex probability densities.    
This property allows VI to converge faster than classical methods, such as, Markov Chain Monte Carlo sampling. 
Conceptually, VI works by choosing a family of probability density functions and then finding the one closest to the actual probability density---often using the Kullback-Leibler (KL) divergence as the optimization metric.
We introduce the Evidence Lower Bound to tractably compute the approximated probability density and we review the ideas behind mean-field variational inference. 
Finally, we discuss the applications of VI to variational auto-encoders (VAE) and VAE-Generative Adversarial Network (VAE-GAN).
With this paper, we aim to explain the concept of VI and assist in future research with this approach.

\end{abstract}
\begin{multicols}{2}


\section{Introduction}

The core principle of Bayesian statistics is to frame inference about unknown variables as a calculation involving a posterior probability density \citep{blei2017variational}.
This property of Bayesian statistics makes inference a recurring problem; especially when the posterior density is difficult to compute \citep{barber2012bayesian}.
Algorithms like the elimination algorithm \citep{cozman2000generalizing}, the message-passing algorithm \citep[belief propagation:][]{barber2012bayesian}, and the junction tree algorithm \citep{madsen1999lazy} have been used to solve exact inference. 
This method involves analytically computing the conditional probability distribution over the variables of interest.
However, the time complexity of exact inference on arbitrary graphical models is NP-hard \citep{dagum1993approximating}.
In the case of large data-sets and complicated posterior probability densities, exact inference algorithms favour accuracy at the cost of speed.
Approximate inference techniques offer an efficient solution by providing an estimate of the actual posterior probability density.

As a solution to approximate inference, various Markov Chain Monte Carlo (MCMC) methods have been extensively studied since the early 1950s. 
The most notable among these methods include the Metropolis-Hastings algorithm \citep{metropolis1953equation} and Gibbs sampling \citep{geman1984stochastic}. 
MCMC techniques have since evolved into an indispensable statistical tool for solving approximate inference. 
However, these methods are slow to converge and do not scale efficiently.

As an alternative to MCMC sampling, variational methods have been used to tractably approximate complicated probability densities. 
In recent years, Variational Inference (VI) \cite[introduced, by][]{jordan1999introduction} has gained popularity in statistical physics \citep{regier2015celeste}, data modeling \citep{tabouy2020variational}, and neural networks \citep{Mackay2015}.
The problem involves using a metric to select a tractable approximation to the posterior probability density \citep{blei2017variational}.
This methodology formulates the statistical inference problem as an optimization problem.
Thus, we get the speed benefits of maximum a posteriori (MAP) estimation \citep{Murphy1991} and can easily scale to large data sets \citep{blei2017variational}.

We organize the paper as follows. 
Section 2 outlines the problem statement and introduces the idea of using KL-divergence, the metric used to measure the information gap between the approximate and the actual posterior probability densities.
Section 3 discusses the concept of evidence lower bound and its importance.
Section 4 introduces the mean-field variational family.
Section 5 applies VI to a toy problem.
Section 6 outlines a few practical applications of VI in the field of deep learning and computer vision. 
Finally, section 7 provides of a summary of the paper.


\section{Problem Statement}
\label{section:prblm_stat}
\begin{figurehere}
    \centering
    \includegraphics[width=0.5\columnwidth]{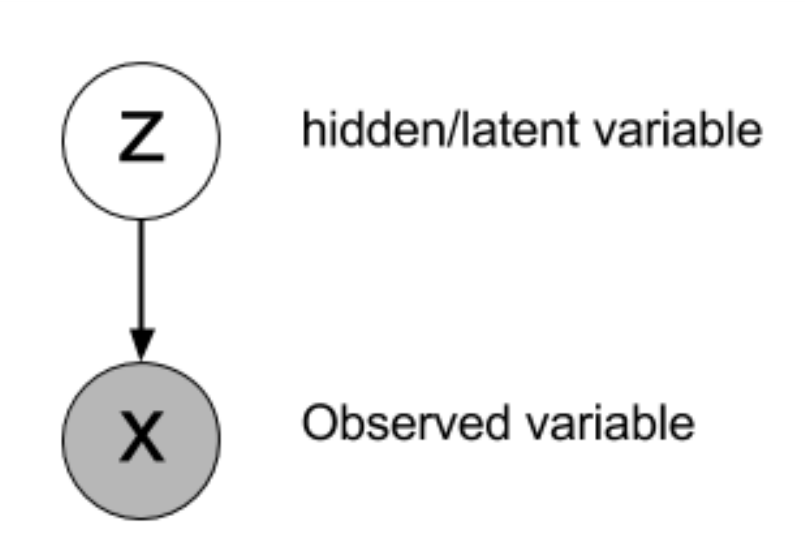}
    \caption{A directed graphical model showing that the observed variable \(X\) is dependent on the latent variable \(Z\).}
    \label{fig:pgm}
\end{figurehere}

\noindent Consider the system of random variables illustrated in Figure \ref{fig:pgm}, where \(X\) and \(Z\) represent the observed variable and the hidden (latent) variable, respectively.
The arrow drawn from \(Z\) to \(X\) represents the conditional probability density \(p(X|Z)\), referred to as the \textit{likelihood}.
From Bayes' theorem we compute the posterior probability density as, 
\begin{align}\label{equation:posterior}
p(Z | X) &= \frac{p(X|Z)p(Z)}{p(X)}.
\end{align}
The marginal, \(p(X)\), can be computed as,
\begin{align}
p(X) &= \int\limits_{z\in Z}p(X|z)p(z) \mathrm{d}z,
\end{align}
where \(z\) is an instance from the sample space of \(Z\). 

\noindent This marginal probability density of observations is the \textit{evidence} and \(p(Z)\) is referred to as the \textit{prior} because it captures the prior information about \(Z\). 
For many models, this evidence integral depends on the selected model and is either unavailable in closed form or requires exponential time to compute \citep{blei2017variational}.

The purpose of VI is to provide an analytical approximation of the posterior probability density \(p(Z|X)\) for statistical inference over the latent variables.
VI enables efficient computation of a lower bound to the marginal probability density, or the evidence.
The idea is that a higher marginal likelihood is indicative of a better fit to the observed data by the chosen statistical model.
Additionally, VI addresses the approximation problem by choosing a probability density function \(q\) for the latent variable \(Z\) from a tractable family \citep{Murphy1991}. 


\subsubsection*{KL-Divergence}
The choice of approximate probability density is done using a metric to measure the difference between it and the actual posterior density \citep{ranganath2014black}.
One popular metric used in VI is the Kullback–Leibler (KL) divergence, suggested by \cite{jordan1999introduction}.
The KL-divergence is the \textit{relative entropy} between two distributions \citep{dembo1991information}.
It is a measure of information that quantifies how similar a probability distribution \(P(X)\) is to a candidate distribution \(Q(X)\) \citep{shlens2014notes}.
The \textit{entropy} is a measure of the mean information or uncertainty of a random variable \(X\) \citep{shannon1948mathematical}, and is defined as
\begin{equation*}
    \mathds{H}(P) = -\sum\limits_{x \in X}P(x)\log P(x),
\end{equation*} 
where \(X\) is sampled from the distribution \(P\).
Subsequently, the KL-divergence, can be expressed as
\begin{equation}
D_{\text{KL}}(P\;\|\;Q) = \sum\limits_{x \in X}P(x)\log\frac{P(x)}{Q(x)},
\label{equation:kld}
\end{equation}
\begin{align*}
    D_{\text{KL}}(P\;\|\;Q) &= -\mathds{H}(P) + \mathds{H}(P, Q),
\end{align*}
where \(\mathds{H}(P, Q)\) is the cross-entropy between the two distributions.
In other words, the KL-divergence is the average extra amount of information required to encode the data using the candidate probability distribution instead of the actual distribution \citep{Murphy1991}.
The KL-divergence is non-negative, non-symmetric and is equal to zero or infinite for two perfectly matching and non-matching distributions, respectively \citep{shlens2014notes}.

For a continuous random variable \(X\), Equation \ref{equation:kld} can be extended to the form,
\begin{equation}
D_{\text{KL}}(P\;\|\;Q) = \int_{-\infty}^{\infty}p(x)\log\frac{p(x)}{q(x)} \mathrm{d}x,
\end{equation}
where \(P\) and \(Q\) are probability distributions of the continuous random variable $X$ and \(p\) and \(q\) represent the probability density functions.

Alternatively, we can express the expectation of the logarithmic difference between the probability densities \(p\) and \(q\) as,
\begin{equation}
D_{\text{KL}}(P\;\|\;Q) = \mathds{E}_{x \sim P(X)} \bigg[ \log \frac{p(x)}{q(x)} \bigg],
\end{equation}
where the random variable \(x\) is sampled from the probability distribution function \(P\) and \(\mathds{E}\) is the expectation function.

As established earlier in this Section, the objective of VI is to select an approximate probability density \(q\) from a family of tractable probability densities \(\mathcal{Q}\). 
Each \(q(Z) \in \mathcal{Q}\) is a candidate approximation of the actual posterior. 
The goal is to find the best candidate, i.e. the one with the minimum KL-divergence \citep{blei2017variational}. 
In our formulation, we assume the approximate probability density is not conditioned on the observed variable. 
Therefore, the inference problem is re-framed as the optimization problem,
\begin{equation}
q^*(Z) = \argmin_{q(Z) \in \mathcal{Q}} D_{\text{KL}}\big(P(Z|X)\;\|\;Q(Z)\big).
\label{equation:forward_kld}
\end{equation}
We optimize Equation \ref{equation:forward_kld} to yield the best approximation \(q^*(.)\) to the actual posterior from the chosen family of densities.
The complexity of the optimization depends our choice for the family of probability densities \citep{Murphy1991,blei2017variational} and, therefore, most researchers choose to use the exponential family---motivated by their conjugate nature.

Computing Equation \ref{equation:forward_kld} is difficult as taking expectations with respect to \(P\) is assumed to be intractable \citep{Murphy1991}. 
Moreover, computing the forward KL-divergence term in Equation \ref{equation:forward_kld} would require us to know the posterior.
An alternative is to use the reverse KL-divergence where the average cross-entropy between the actual posterior and our approximation is computed by taking expectations, with respect to the variational distribution.
Hence, the optimization problem in Equation \ref{equation:forward_kld} can be re-formulated as,
\begin{equation}
q^*(Z) = \arg \min_{q(Z) \in \mathcal{Q}} D_{\text{KL}}(Q(Z)\;\|\;P(Z|X)).
\label{equation:rev_kld}
\end{equation}
Since \(q(Z)\) is selected from a tractable family of probability densities, computing expectations with respect to \(q\) is also tractable. 


\subsubsection*{Forward vs. Reverse KL}
Let \(P\) and \(Q\) be two distributions with probability density functions \(p\) and \(q\), where \(q\) is an approximation of \(p\).
As stated earlier, KL-divergence is non-symmetric \citep{shlens2014notes}, i.e., 
\begin{equation*}
    D_{\text{KL}}(P\;\|\;Q) \neq D_{\text{KL}}(Q\;\|\;P),
\end{equation*}
as such, minimizing the forward KL-divergence, \(D_{\text{KL}}(P\;\|\;Q)\), yields different results than minimizing the reverse KL-divergence, \(D_{\text{KL}}(Q\;\|\;P)\).

The forward KL-divergence is also known as the M-projection or moment projection, \citep{Murphy1991} and is defined as,
\begin{equation*}
    D_{\text{KL}}(P\;\|\;Q) = \mathds{E}_{x \sim P(X)}\bigg[\log \frac{p(x)}{q(x)}\bigg].
\end{equation*}
This will be large wherever the approximation fails to \textit{cover up} the actual probability distribution \citep{Murphy1991}; i.e. 
\begin{equation*}
    \lim_{q(x)\to 0} \frac{p(x)}{q(x)} \to \infty, \; \text{where} \; p(x) > 0.
\end{equation*}
So, if \(p(x) > 0\), we must choose a probability density to ensure that \(q(x) > 0\) \citep{Murphy1991}. 
This particular case of optimizing is \textit{zero avoiding} and can intuitively be interpreted as \(q\) over-estimating \(p\).

\begin{figure*}
    \centering
    \includegraphics[width=\textwidth]{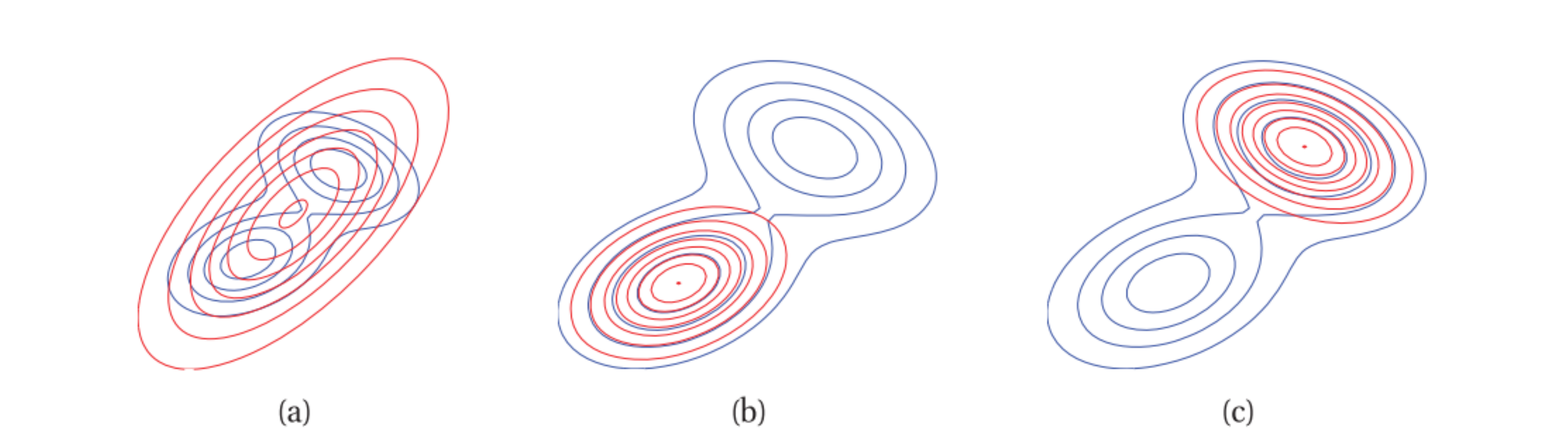}
    \caption{Figure illustrating forward vs reverse KL-divergence on a bimodal distribution. The blue and the red contours represent the actual probability density, \(p\), and the unimodal approximation, \(q\), respectively. The left panel shows the forward KL-divergence minimization where \(q\) tends to \textit{cover} \(p\). The centre and the right panels show the reverse KL-divergence minimization where \(q\) locks on to one of the two modes.}
    \label{fig:fwd_vs_rev}
\end{figure*}

The reverse KL-divergence is also known as the I-projection or information projection, \citep{Murphy1991} and is defined as,
\begin{equation*}
    D_{\text{KL}}(Q\;\|\;P) = \mathds{E}_{x \sim Q(X)}\biggl[\log \frac{q(x)}{p(x)}\biggr],
\end{equation*}
where, 
\begin{equation*}
    \lim_{p(x)\to 0} \frac{q(x)}{p(x)} \to \infty \; \text{where} \; q(x) > 0.
\end{equation*}
The limit indicates the need to force \(q(x) = 0\) wherever \(p(x) = 0\), otherwise the KL-divergence would be very large. 
This is \textit{zero forcing} \citep{Murphy1991} and can be interpreted as \(q\) under-estimating \(p\). 
The difference between the two methods is illustrated in Figure \ref{fig:fwd_vs_rev}; based on Figure 21.1 from \cite{Murphy1991}.

\section{ELBO: Evidence Lower Bound}
\label{section:ELBO}
As mentioned in Section \ref{section:prblm_stat}, we select a probability density from a tractable family which has the lowest KL-divergence from the actual posterior density.
Therefore, inference amounts to solving the optimization problem defined in Equation \ref{equation:rev_kld}.
However, optimizing Equation \ref{equation:rev_kld} is still not tractable because we are required to compute the evidence function. 
The KL-divergence objective function from Equation \ref{equation:rev_kld} can be written as,
\begin{align}
\mathcal{D} &= \mathds{E}[\log q(z)] - \mathds{E}[\log p(z|x)],
\end{align}
where,
\begin{equation}
\mathcal{D} = D_{\text{KL}}(Q(Z)\;\|\;P(Z|X)),
\end{equation}
and all expectations are taken by sampling \(z\) from \(Q(Z)\).
We now expand the conditional probability density \(p(z|x)\) using Equation \ref{equation:posterior}, giving us,
\begin{align}
\mathcal{D} &= \mathds{E}[\log q(z)] - \mathds{E}[\log p(z,x)] + \mathds{E}[\log p(x)].
\label{equation:kld_log_p_x}
\end{align}
Since all expectations are under \(Q(Z)\), \(\mathds{E}[\log p(x)]\) is the constant \(\log p(x)\).
Therefore, we can re-write Equation \ref{equation:kld_log_p_x} as, 
\begin{align}
\mathcal{D} = \mathds{E}[\log q(z)] - \mathds{E}[\log p(z,x)] + \log p(x).
\label{equation:kld_expand}
\end{align}
The KL-divergence cannot be computed directly as it depends on the evidence.
Therefore, we must optimize an alternative objective function that is equivalent to \(D_{\text{KL}}\) up to an added constant,
\begin{align}
-\mathcal{D} + \log p(x) &= \mathds{E}[\log p(z, x)] - \mathds{E}[\log q(z)] ,\nonumber \\
\text{ELBO}(Q) &= \mathds{E}[\log p(z, x)] - \mathds{E}[\log q(z)],
\label{equation:elbo}
\end{align}
where the term ELBO is an abbreviation for evidence lower bound.
The ELBO the sum of the negative KL-divergence and the constant term \(\log p(x)\). 
Maximizing the ELBO is equivalent to minimizing the KL-divergence \citep{blei2017variational}.

An intuitive explanation of the ELBO can be derived by re-arranging the terms of Equation \ref{equation:elbo}, as
\begin{align}
\text{ELBO}(Q) &= \mathds{E}[\log p(z, x)] - \mathds{E}[\log q(z)],\nonumber \\
&= \mathds{E}[\log p(x|z)] + \mathds{E}[\log p(z) - \log q(z)], \nonumber \\
&= \mathds{E}[\log p(x|z)] - D_{\text{KL}}(Q(Z)\;\|\;P(Z)).
\label{equation:elbo_final}
\end{align}
Thus, the ELBO is the sum of the expected log likelihood of the data and the KL-divergence between the prior and approximated posterior probability density. 
The expected log likelihood describes how well the chosen statistical model fits the data.
The KL-divergence encourages the variational probability density to be close to the actual prior. 
Thus, the ELBO can be seen as a regularised fit to the data.

The ELBO lower-bounds the (log) evidence, \(\log p(x)\). 
This property was explored by \cite{jordan1999introduction}, where the authors used Jensen's inequality \citep{klarivcicsome} to derive the relationship between the ELBO and the evidence function. The derivation is as follows:
\begin{align}
\log p(x) &= \log \int\limits_{z\in Z}p(x,z)\mathrm{d}z,\nonumber \\
&= \log \int\limits_{z\in Z}p(x,z)\frac{q(z)}{q(z)}\mathrm{d}z, \nonumber \\
&= \log \mathds{E}_{z \sim Q(z)}\biggl[\frac{p(x,z)}{q(z)}\biggr], \nonumber \\
&\geq \mathds{E}_{z \sim Q(z)}[\log p(x,z)] - \mathds{E}_{z \sim Q(z)}[\log q(z)] \nonumber, \\
&\geq \text{ELBO}(Q).
\end{align}
This relationship between the ELBO and \(\log p(x)\) has motivated researchers to use the variational lower bound as the criterion for model selection. 
This bound serves as a good approximation to the marginal likelihood; providing a basis for model selection.
Applications of VI for model selection have been explored in a wide variety of tasks such as for mixture models \citep{mcgrory2007variational}, cross-validation mode selection \citep{nott2012regression} and in a more general setting by \cite{bernardo2003variational}. 


\section{Mean field variational family}
\label{section:mean_field}
We briefly, introduce the mean field variational family for VI, where the latent variables are assumed to be mutually independent---each governed by a distinct factor in the variational probability density \citep[see][for a more detailed explanation]{10.5555/1162264}.
This assumption greatly simplifies the complexity of the optimization process.
A generic member of the mean field variational family is
\begin{equation}
q(Z|X) = \prod\limits_{j=1}^{m}q_{j}(Z_{j}),
\label{equation:mean_field}
\end{equation}
where \(m\) is the number of latent variables.
The observed data \(X\) does not appear in Equation \ref{equation:mean_field}, therefore any probability density from this variational family is not a model of the data.
Instead, it is the ELBO, and the corresponding KL-divergence minimization problem, which connects the fitted variational probability density to the data and model \citep{blei2017variational, Murphy1991}.


\section{A toy problem}
In this section, we explore, in detail, how VI can be used to approximate a mixture of Gaussians.
Consider a distribution of \(N\) real-valued data-points \(x=x_{1},
x_{2}, ..., x_{N}\) sampled from a mixture of \(K\) univariate Gaussians with means \(\mu_j = \mu_{1}, \mu_{2},..., \mu_{K}\). 
We assume that the variance of the mean's prior is a fixed hyperparameter \(\sigma^{2}\) while the observation variance is one.
For this problem, we define a single data-point as,
\begin{equation}
    x_{i} \sim \mathcal{N}(c_{i}^{T}\mu, 1) \text{  for  } i = 1,2,...N,
    \label{equation:true_x}
\end{equation}
which is drawn from a distributions with mean,
\begin{equation}
    \mu_{j} \sim \mathcal{N}(0, \sigma^{2}) \text{  for  } j = 1,2,...,K,
\end{equation}
and is assigned to a cluster using,
\begin{equation}
    c_{i} \sim \mathcal{U}(K) \text{  for  } i = 1,2,...,N,
\end{equation}
where \(c_{i}\) is a one-hot vector of \(K\)-dimensions, and with latent variables \(\mu\) and \(c\).
In our case, the one-hot vector is a \(K\)-dimensional binary vector where each dimension represents a cluster.
A data-point belonging to the \(l\)-th cluster will be represented by a value of one in the \(l\)-th dimension of the one-hot vector, while the \((K-1)\) remaining dimensions will have a value of zero.

We assume the approximate posterior probability density to be from the mean-field variational family (Section \ref{section:mean_field}). 
Thus, the variational parameterization is given by,
\begin{equation}
    q(\mu, c) = \prod_{j=1}^{K}q(\mu_{j}; m_{j}, s_{j}^2) \prod_{i=1}^{N}q(c_{i}; \phi_{i}).
    \label{equation:mf_gmm_init}
\end{equation}
Each latent variable is governed by it's own variational factor \citep{blei2017variational}. 
Here, the mixture components are Gaussian with variational parameters (mean \(m_{k}\) and variance \(s_{k}^2\)) specific to the \(k\)-th cluster. 
The cluster assignments are categorical with variational parameters (\(K\)-dimensional cluster probabilities \(\phi_{i}\) vector) specific to the \(i\)-th data point.

The definition of ELBO from Equation \ref{equation:elbo} applied to this specific case is,
\begin{align}
    \text{ELBO}_{m, s^{2}, \phi} &= \mathds{E}[\log p(x, \mu, c)] \nonumber\\
    &- \mathds{E}[\log q(\mu, c)],
    \label{equation:elbo_gmm__init}
\end{align}
\begin{equation*}
    \text{ELBO}_{m, s^{2}, \phi} = \text{ELBO}(m, s^{2}, \phi).
\end{equation*}
We maximize the ELBO to derive the optimal values of the variational parameters (see Appendix \ref{section:appendix_A}). 
The optimal values for \(m\), \(s\) and \(\phi\) are given by,
\begin{align}
    m_{j}^* &= \frac{\sum_{i} \phi_{ij}x_{i}}{\frac{1}{\sigma^{2}} + \sum_{i}\phi_{ij}}, \\
    (s_{j}^{2})^* &= \frac{1}{\frac{1}{\sigma^{2}} + \sum_{i}\phi_{ij}}, \\  
    \phi_{ij}^* &\propto e^{-\frac{1}{2}(m_{j}^{2}+s_{j}^{2}) + x_{i}m_{j}}.
\end{align}
We employ the Coordinate Ascent VI (CAVI) algorithm to optimize the ELBO. 
Algorithm \ref{algoritm:2} \citep{blei2017variational} describes the steps to optimize the ELBO using CAVI.

\begin{figure*}
    \centering
    \includegraphics[width=\columnwidth]{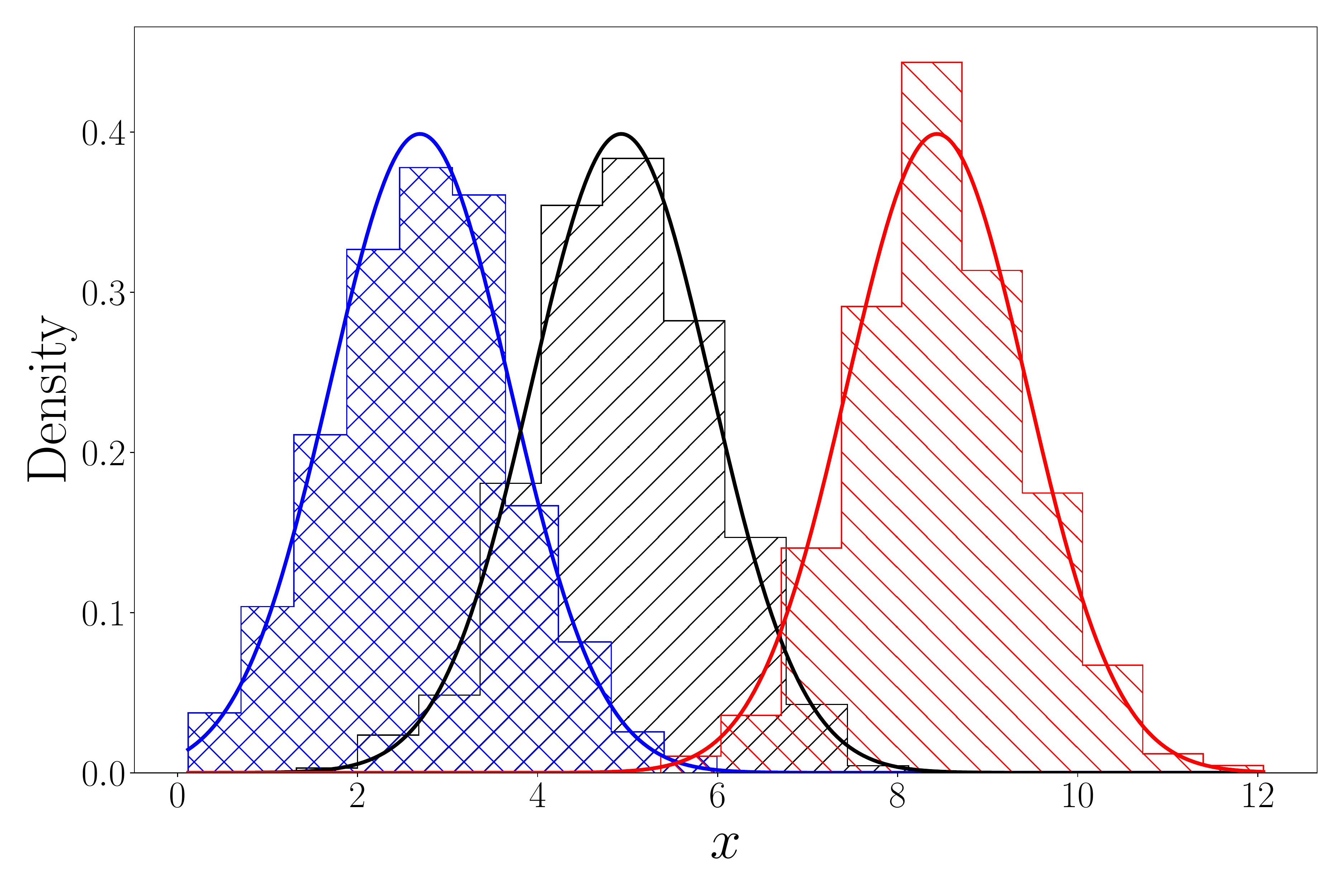}
    \includegraphics[width=\columnwidth]{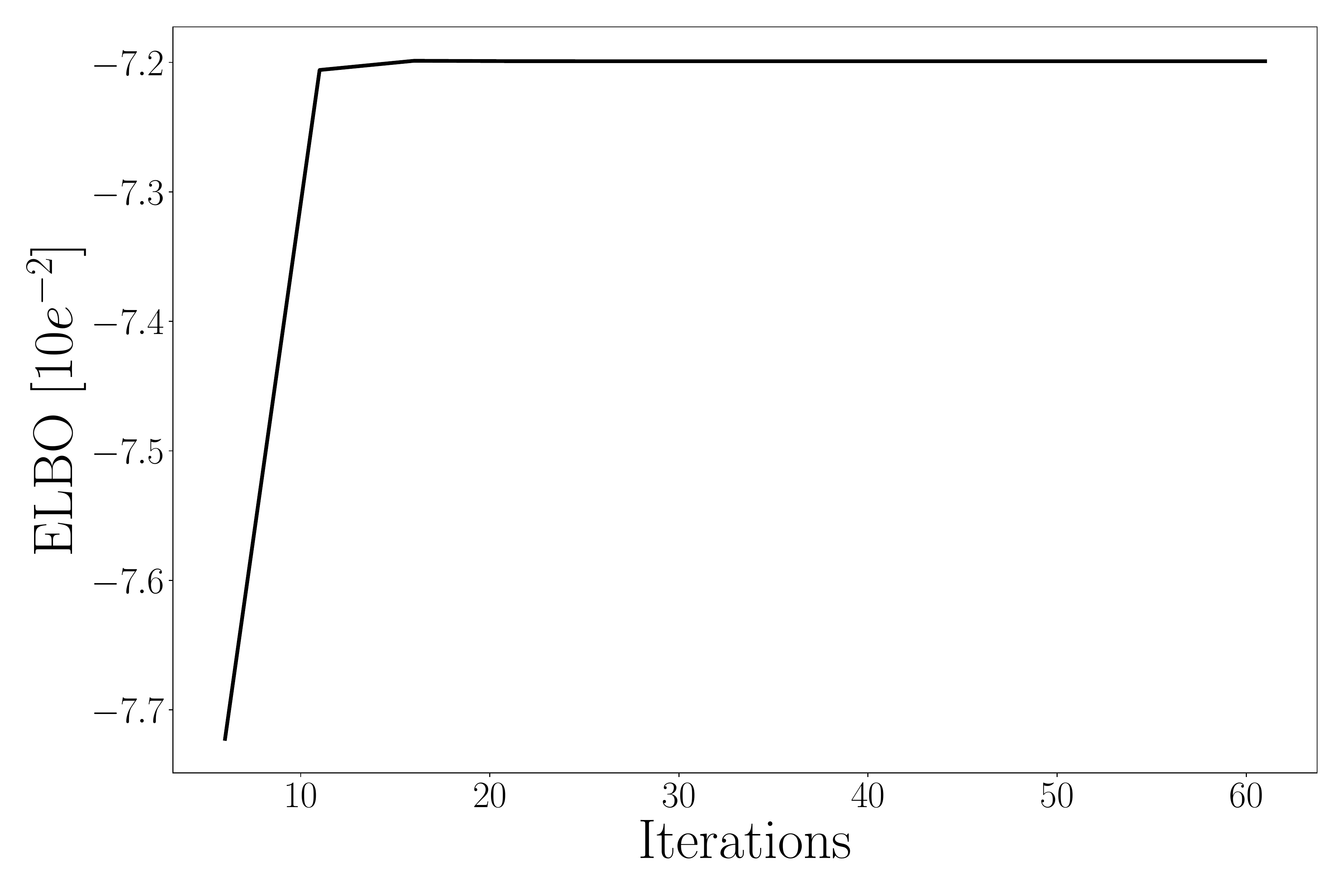}
    \caption{Left: a histogram showing the distribution of data sampled from the three univariate Gaussians. The curved lines indicate the fit by maximising the ELBO.
    Right: an illustration of the convergence of ELBO using CAVI.}
    \label{fig:gmm_fit}
\end{figure*}

For our experimental setup, we select \(K=3\), i.e., a mixture of three univariate Gaussians. We generate the data by randomly sampling 1000 data-points for each of the three Gaussians.
The left panel in Figure \ref{fig:gmm_fit} illustrates how the algorithm approximates the parameters of each individual Gaussian by maximizing the ELBO. 
We select a maximum iteration of 1000 steps, however, the ELBO converges by iteration 60 as illustrated in the right panel of Figure \ref{fig:gmm_fit}.

\begin{algorithm}[H]
\SetAlgoLined
\DontPrintSemicolon
\KwData{Data \(x_{1:n}\), \(K\) mixture components and prior variance of component means \(\sigma^{2}\)}
\KwResult{Variational densities \(q(\mu_{j}; m_{j}, s_{j^{2}})\) and \(q(c_{i}; \phi_{i})\)}
$ m = m_{1:K}, s^{2} = s_{1:K}^{2}, \phi = \phi_{1:N} \gets$ initialize variational parameters \\
\While{the ELBO has not converged}{
    \For{$i \in  {1, ..., N}$}{
        Set $\phi_{ij} \propto \exp \big[{-\frac{1}{2}(m_{j}^{2}+s_{j}^{2}) + x_{i}m_{j}}\big]$}
    \For{$j \in {1,..., K}$}{
        Set $m_{j} \longleftarrow \frac{\sum_{i} \phi_{ij}x_{i}}{1 / \sigma^{2} + \sum_{i}\phi_{ij}}$ \\
        Set $s_{j}^{2} \longleftarrow \frac{1}{1 / \sigma^{2} + \sum_{i}\phi_{ij}}$
    }
    Compute $\text{ELBO}(m, s^{2}, \phi)$
    }
\Return{$q(m, s^2, \phi)$}
\caption{CAVI for a Gaussian mixture model}
\label{algoritm:2}
\end{algorithm}

\section{Applications}
We have established the optimization process for VI, next we look at some applications in generative modelling.


\subsection{VAE: Variational Auto-Encoder}\label{section:vae}
An auto-encoder is a neural network that aims to learn or encode a low-dimensional representation for high-dimensional data, e.g. images.
Different variants of auto-encoders exist that aim to learn meaningful representations of high-dimensional data.
One such variant is the Variational Auto-Encoder (VAE).
Introduced by \cite{kingma2013auto}, the VAE is a statistical model which is essentially a stochastic variational inference algorithm.
The VAE uses the concept of variational inference to compress the high-dimensional data into a latent vector while assuming a multi-variate distribution as a prior for the same latent vector.
The statistical model uses gradient backpropagation to approximate the posterior distribution for the latent vector.
For large data sets we update the VAE's parameters using small mini-batches or even single data points.

Since their inception, VAEs have been widely used for generative modelling.
They are easy to implement, converge faster than MCMC methods, and scale efficiently to large data sets; this makes them ideal for generative modelling of image data. 
However, the images they generate tend to have reduced quality compared to the input images. 
This is the effect of minimizing the reverse KL-divergence, which results in the approximate distribution being locked to one of the modes---as explained in Section \ref{section:ELBO}. 

\cite{kingma2013auto} introduce the stochastic variational inference algorithm (VAE) that reparameterizes the variational lower bound, yielding a lower bound estimator that can be optimized using standard stochastic gradient methods \citep{kingma2013auto}.

\begin{figurehere}
    \centering
    \includegraphics[width=0.6\columnwidth]{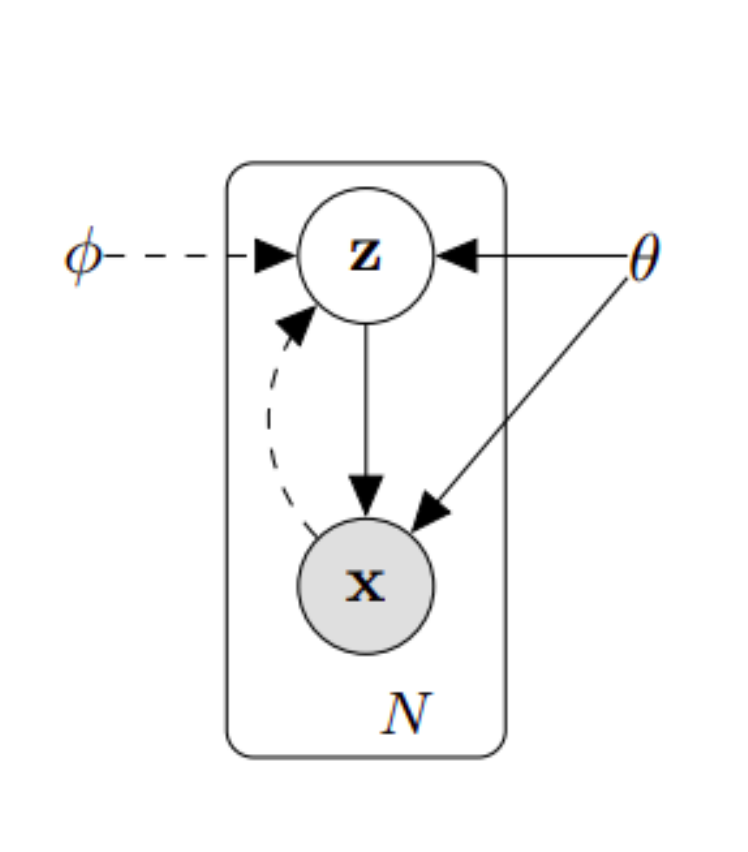}
    \caption{
    Illustration of the type of directed graphical model under consideration with \(N\) observed data-points. Solid lines denote the generative model, dashed lines denote the variational approximation to the intractable posterior density.
    The variational parameters \(\phi\) are learned jointly with the generative model parameters \(\theta\) \citep{kingma2013auto}.}
\end{figurehere}

The recognition model, \(q_{\phi}(Z|X)\), can be interpreted as a probabilistic encoder, since given a data point \(x\) the encoder produces a latent vector \(z\).
This latent vector is then used to generate a sample from the likelihood density, \(p_{\theta}(X|Z)\), and, therefore, the generative model can be interpreted as a probabilistic decoder \citep{kingma2013auto}.
The probability density \(q_{\phi}(Z|X)\) serves as an approximation of the actual posterior probability density \(p_{\theta}(Z|X)\).
Using a mini-batch of data points sampled from \(X\), the encoder transforms these data points into the latent space, \(Z\), which the decoder uses to generate a the samples in \(X\). 

As established in Equation \ref{equation:elbo_final}, maximizing the ELBO is equivalent to minimizing the KL-divergence \citep{blei2017variational}. 
In order to jointly optimize the recognition model and the generative model on mini-batches of data, we differentiate and optimize the lower bound with respect to both the variational and the generative parameters, $\phi$ and $\theta$. 
In this case Equation \ref{equation:mean_field} can be rewritten as,
\begin{align}
\mathcal{D}_{Z, X, \phi,\theta} &= D_{\text{KL}}(Q_{\phi}(Z|X)\;\|\;P_{\theta}(Z)), \nonumber\\
\mathcal{L}(\phi, \theta; x) &= -\mathcal{D}_{Z, X, \phi,\theta} + \mathds{E}_{z \sim Q_{\phi}(Z|X)}[\log p_{\theta}(x|z)].
\label{equation:vae}
\end{align}
If we look closely at the terms on the right-hand-side of Equation \ref{equation:vae}, we can see the connection to auto-encoders, where the second term is the expected negative reconstruction error and the KL-divergence term can be interpreted as a regularization term.
In order to optimize Equation \ref{equation:vae} using standard gradient-based techniques, \cite{kingma2013auto} introduce the Auto-Encoding Variational Bayes (AEVB) algorithm to efficiently compute the gradient of the ELBO from Equation \ref{equation:vae}.
For a chosen approximate posterior \(q_{\phi}(z|x)\), we re-parameterize the random variable \(z = q_{\phi}(z|x)\) with a differentiable transformation \(g_{\phi}(\epsilon, x)\) of a noise variable \(\epsilon\), such that,
\begin{align*}
    z &= g_{\phi}(\epsilon, x), \\
    \epsilon &\sim P(\epsilon).
\end{align*}
This re-parameterization allows us to form Monte Carlo estimates of expectations of the function \(\log p_{\theta}(x|z)\).
Forming Monte Carlo estimates enables us to numerically evaluate the expectation,
\begin{equation*}
    \mathds{E}_{z \sim Q_{\phi}(Z|X)}[\log p_{\theta}(x|z)].
\end{equation*}
We draw independent samples, $z^{i}$, from the variational distribution, \(Q_{\phi}(Z|X)\), and then compute the average of the function evaluated at these samples \citep{mohamed2020monte}.
Therefore, the Monte Carlo estimates of the expectation of the function \(\log p_{\theta}(x^{(i)}|z)\) when \(z \sim q_{\phi}(z|x^{(i)})\) is as follows, 
\begin{align*}
    \mathds{E}_{z \sim Q_{\phi}(Z|x^{(i)})} \simeq \frac{1}{L}\sum_{l=1}^{L}\log p_{\theta}(x^{(i)}|z^{(i,l)}),
\end{align*}
where, \(\epsilon^{(l)} \sim P(\epsilon)\) and \(L\) is the number of samples per data point.

Applying the above re-parameterization to the variational lower-bound of Equation \ref{equation:vae} we can re-formulate the ELBO to a Stochastic Gradient Variational Bayes (SGVB) estimator \(\mathcal{L}(\phi, \theta; x^{(i)})\), as
\begin{equation}
\mathcal{L}_{\phi,\theta,x^{(i)}} \simeq -\mathcal{D}_{Z, X, \phi,\theta} +  \frac{1}{L}\sum_{l=1}^{L}\log p_{\theta}(x^{(i)}| z^{(i,l)})],
\label{equation:vae_reparam}
\end{equation}
\begin{equation*}
\mathcal{L}_{\phi,\theta,x^{(i)}} = \mathcal{L}(\phi, \theta; x^{(i)}).
\end{equation*}
We re-parameterize the variational lower bound in terms of a deterministic random variable which enables us to use gradient based optimizers on mini-batches of data. 
This further enables the optimization of the parameters of the distribution while still maintaining the ability to randomly sample from that distribution \citep{doersch2016tutorial}.

\begin{figurehere}
    \centering
    \includegraphics[width=0.8\columnwidth]{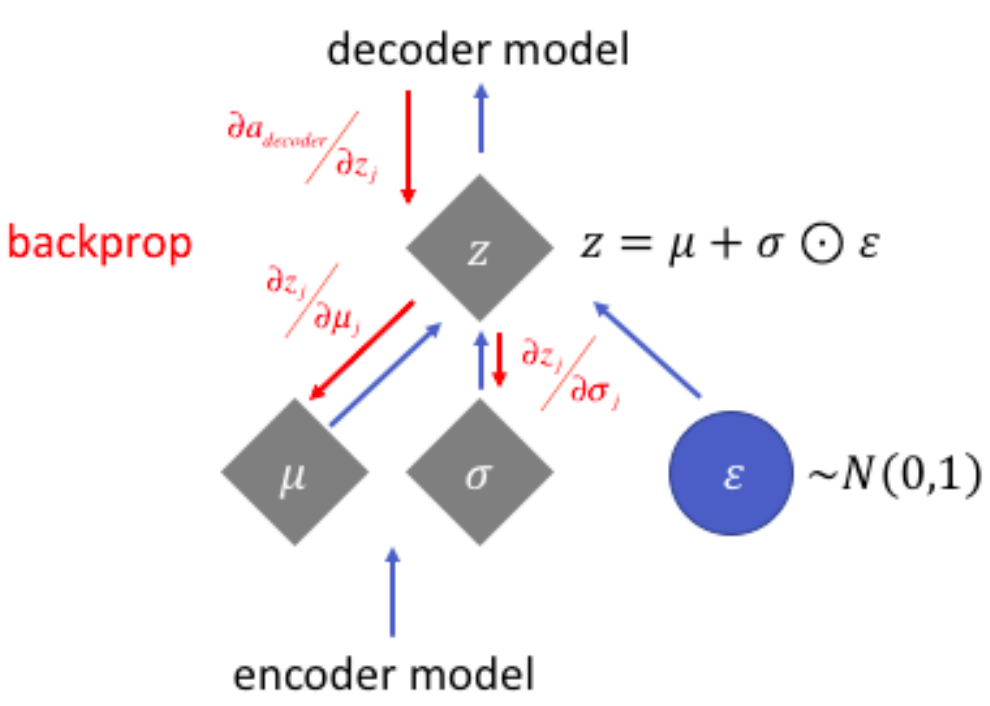}
    \caption{The learning process in a typical VAE using gradient back-propagation.}
\end{figurehere}

\begin{figure*}
    \centering
    \includegraphics[width=0.6\textwidth]{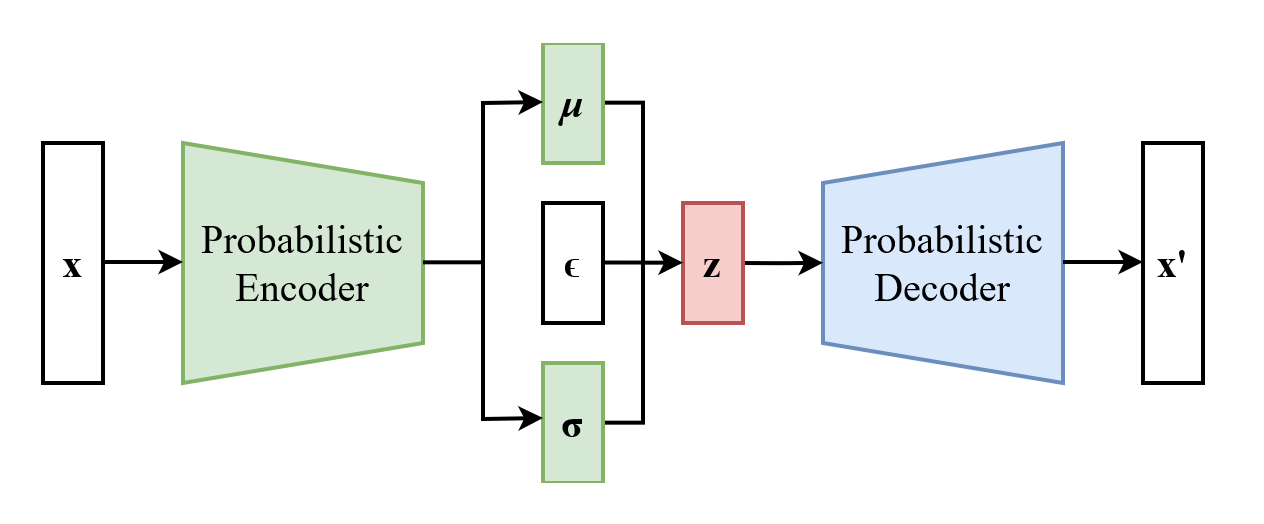}
    \caption{A diagram showing a variational autoencoder model.}
\end{figure*}

In order to simplify the calculations, we assume the variational approximate posterior to be a multi-variate Gaussian with diagonal co-variance structure \citep{kingma2013auto}. 
As for the prior we assume a multivariate Gaussian \(\mathcal{N}(z; 0,\textbf{I})\) where,
\begin{align*}
\log q_{\phi}(z | x^{(i)}) &= \log \mathcal{N}(z; \mu^{(i)}, (\sigma^{(i)})^{2}\textbf{I}),\\
p_{\theta}(z) &= \mathcal{N}(z; 0,\textbf{I}), \\
z^{(i)} &= \mu^{(i)} + \sigma^{(i)} \odot \epsilon, \\
\epsilon &\sim \mathcal{N}(0, \textbf{I}).
\end{align*}
Using the above parameterization, the KL-divergence term in Equation \ref{equation:vae_reparam} can be derived as Equation \ref{equation:box} (as shown in Box 1).
Subsequently, Equation \ref{equation:vae_reparam} can be used to define the loss function for the VAE framework at \(x^{(i)}\), as
\begin{align}
    \mathcal{L}_{\phi, \theta; x^{(i)}} &\simeq \frac{1}{2}\sum_{j=1}^{J}\Big[ 1 + \log(   (\sigma_{j}^{(i)})^{2}) - (\mu_{j}^{i})^{2} - (\sigma_{j}^{(i)})^{2}\Big] \nonumber\\
     &+ \frac{1}{L}\sum_{l=1}^{L}\log p_{\theta}(x^{(i)}| z^{(i, l)}),
     \label{equation:final_vae_loss}
\end{align}
where \(J\) is the dimensionality of \(z\).

A variant of the VAE framework, \(\beta\)-VAE, adds an extra hyperparameter to the VAE objective which constricts the effective encoding capacity of the latent space.
The \(\beta\)-VAE training objective is,
\begin{align*}
    \mathcal{L}(\phi, \theta; x) &= -\beta \mathcal{D}_{Z, X, \phi,\theta} \\
    &+ \mathds{E}_{z \sim Q_{\phi}(Z|X)}[\log p_{\theta}(x|z)],
\end{align*}
where \(\beta = 1\) corresponds to the original VAE formulation of \cite{kingma2013auto}.
This constriction encourages the latent representation to be more factorised.
However, this can lead to even worse reconstruction quality as compared to the standard VAE framework. 
This is caused by a trade-off introduced by the modified training objective that punishes reconstruction quality in order to encourage disentanglement between the latent representations \citep{burgess2018understanding}. 
Varying \(\beta\) during training encourages the model to learn different latent representations of the data. 
A high value of \(\beta\) encourages disentanglement in the latent space but at the cost of reduced reconstruction quality.
To mitigate this reconstruction problem, the authors introduce a capacity control parameter \(\mathcal{C}\).
Increasing \(\mathcal{C}\) from zero to a value large enough produces good quality reconstructions during training \citep[][]{burgess2018understanding}.
The modified \(\beta\)-VAE training objective is given by,
\begin{align*}
    \mathcal{L}(\phi, \theta; x) &= -\beta | \mathcal{D}_{Z, X, \phi,\theta} - \mathcal{C}|\\ 
    &+ \mathds{E}_{z \sim Q_{\phi}(Z|X)}[\log p_{\theta}(x|z)].
\end{align*}

\end{multicols}
\begin{empheq}[box={\textfbox[Box 1: Derivation of the KL-divergence.]}]{align}
    -\mathcal{D}_{Z, x^{(i)}, \phi,\theta} &= D_{\text{KL}}(Q_{\phi}(Z|x^{(i)})\;\|\;P_{\theta}(Z)) \nonumber \\
    &= -\mathds{E}_{z \sim Q_{\phi}(Z|X)}\biggl[ \log \frac{q_{\phi}(z|x^{(i)})}{p_{\theta}(z)}\biggr] \nonumber \\
    &= -\int q_{\phi}(z|x^{(i)}) \biggl[ \log \frac{q_{\phi}(z|x^{(i)})}{p_{\theta}(z)}\biggr] \mathrm{d}z \nonumber \\
    &= \int q_{\phi}(z|x^{(i)}) \log p_{\theta}(z) dz
    -\int q_{\phi}(z|x^{(i)}) \log q_{\phi}(z|x^{(i)}) \mathrm{d}z \nonumber \\
    &= \int \mathcal{N}\big[z; \mu^{(i)}, (\sigma^{(i)})^{2}\textbf{I}\big] \log \mathcal{N}\big[z; 0, \textbf{I}\big] \mathrm{d}z
    -\int \mathcal{N}\big[z; \mu^{(i)}, (\sigma^{(i)})^{2}\textbf{I} \big] \log \mathcal{N} \big[ z; \mu^{(i)}, (\sigma^{(i)})^{2}\textbf{I}\big] \mathrm{d}z \nonumber \\
    &= \frac{1}{2}\biggl[J\log (2\pi) + \sum_{j=1}^{J}(1 + \log (\sigma_{j}^{(i)})^{2} )
    - J\log (2\pi) - \sum_{j=1}^{J}((\mu_{j}^{(i)})^2 + (\sigma_{j}^{(i)})^{2})\biggr] \nonumber \\
    &= \frac{1}{2}\sum_{j=1}^{J}\biggl[ 1 + \log (\sigma_{j}^{(i)})^{2} - (\mu_{j}^{(i)})^2 -  (\sigma_{j}^{(i)})^{2}\biggr]
     \label{equation:box}
\end{empheq}
\begin{multicols}{2}


\subsection{GAN: Generative Adversarial Network}
Generative adversarial networks (GANs), introduced by \cite{goodfellow2014generative}, are deep-learning based generative models that are extensively used to create realistic data samples across a range of problems---most notably in computer vision. 

A typical GAN framework involves simultaneously training two models; a generator and a discriminator.
The generator tries to capture the data distribution by mapping a latent vector to a data-point, thereby generating new samples with similar statistical properties as the training data.
The discriminator aims to estimate the likelihood of a sample being drawn from the training data or created by the generator
\citep{goodfellow2014generative}. 
We can train a GAN by minimizing the objective function,
\begin{align}
\min_{\text{Gen}} \max_{\text{Dis}} \mathcal{L}_{\text{GAN}} &= \mathds{E}_{x \sim P_{\text{data}}(X)}[\log (\text{Dis}(x))] \nonumber \\
&+ \mathds{E}_{z \sim P_{z}(Z)}[\log (1 - \text{Dis}(\text{Gen}(z)))],
\end{align}
where the generator, $\text{Gen}(z)$, takes a sample from the latent distribution, $P_z(Z)$, and creates a new data-point while the discriminator, $\text{Dis}(x)$, takes a data-point from both the real distribution, $P_{\text{data}}(X)$, and the new data-point $\text{Gen}(z)$ and assigns probabilities to both.
To minimize the objective, the discriminator will try to assign probabilities close to zero and one for data sampled from the real distribution and anything created by the generator, respectively.
The GAN's objective is to train the discriminator to efficiently discriminate between real and generated data while encouraging the generator to reproduce the true data distribution \citep{goodfellow2014generative, larsen2016autoencoding}.
There is a unique solution where the generator successfully recovers the training data distribution. 
At the same time, the discriminator ends up assigning equal probability to samples from the training data and the generator. 


\subsection{VAE-GAN}
Different variants of the original GAN framework have evolved since its inception, such as the VAE-GAN introduced by \cite{larsen2016autoencoding}.
This approach uses the learned feature representations in the GAN discriminator as a basis for the VAE reconstruction objective.

\cite{larsen2016autoencoding} show that unsupervised training, like that of a GAN, can result in the latent image representation with disentangled factors of variation \citep{bengio2013representation}. 
This means that the model learns an embedding space with, abstract, high-level visual features which can be modified using simple arithmetic \citep{larsen2016autoencoding}.

\begin{figurehere}
    \centering
    \includegraphics[width=\columnwidth]{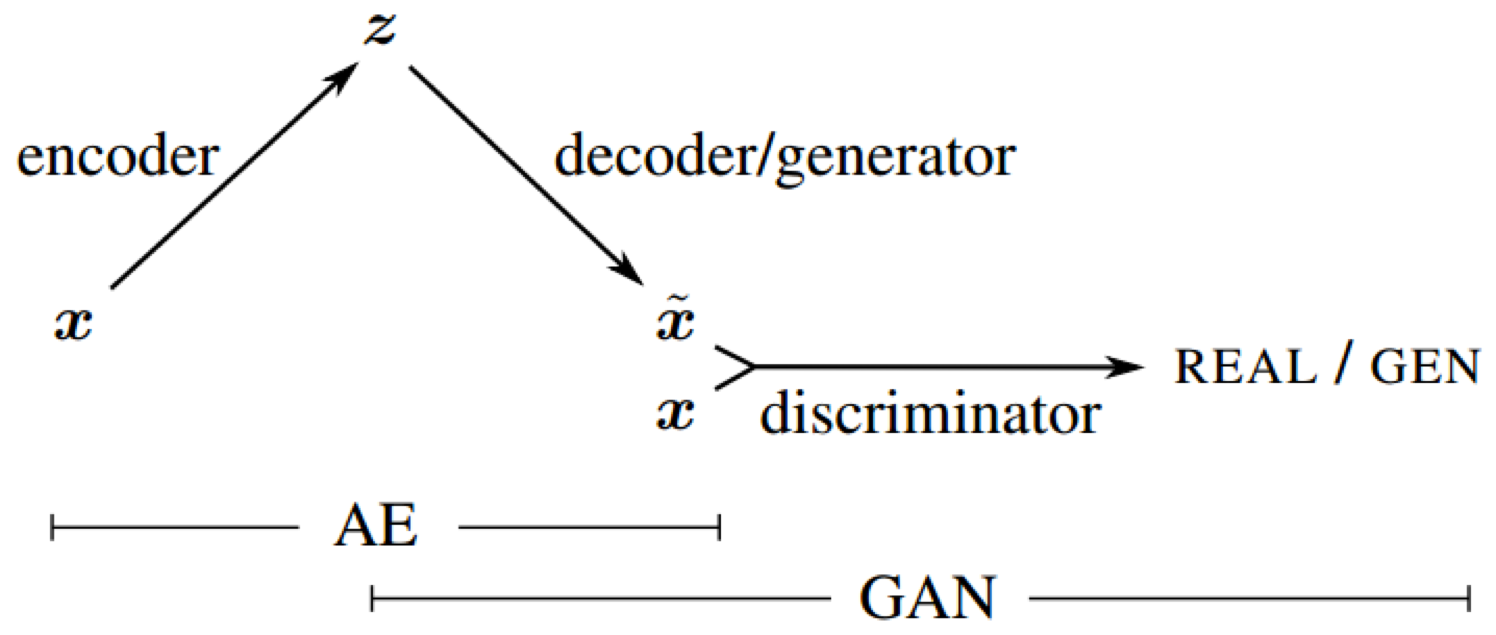}
    \caption{Diagram of the VAE-GAN framework \cite{larsen2016autoencoding}.}
\end{figurehere}

As we see in Section \ref{section:vae}, the VAE consists of an encoder and a decoder given by,
\begin{align*}
    z &= \text{\text{Enc}}(x) = q(z|x), \\
    \Tilde{x} &=\text{Dec}(z) = p(x|z),
\end{align*}
where \(z\) is the latent representation of a data sample \(x\) taken from marginal likelihood distribution.
The encoder, $\text{Enc}(x)$, takes a data-sample, \(x\), and approximates the posterior density \(q(z|x)\).
Whereas the decoder, $\text{Dec}(z)$, takes a sample from the latent space, \(z\), and generates a sample from the likelihood density \(p(x|z)\). 
\cite{larsen2016autoencoding} define \(\mathcal{L}_{\text{VAE}}\) as the negative of training objective of a vanilla VAE (given by Equation \ref{equation:vae}) and provide the objective function, 
\begin{align}
    \mathcal{L}_{\text{VAE}} & = -\mathcal{L}(\phi, \theta; x), \nonumber \\
    &= D_{\text{KL}}(Q(Z|X)\;\|\;P(Z)) \nonumber\\
    &-\mathds{E}_{z \sim Q(Z|X)}[\log p(x|z)].
    \label{equation:l_vae}
\end{align}
The authors go on to describe the terms of Equation \ref{equation:l_vae} as, 
\begin{align*}
    \mathcal{L}_{\text{llike}}^{\text{pixel}} &= -\mathds{E}_{z \sim Q(Z|X)}[\log p(x|z)], \\
    \mathcal{L}_{\text{prior}} &= D_{\text{KL}}(Q(Z|X)\;\|\;P(Z)),
\end{align*}
where \(\mathcal{L}_{\text{llike}}^{\text{pixel}}\) is the negative expected log likelihood and \(\mathcal{L}_{\text{prior}}\) is the KL-divergence between the approximated posterior density and the prior on the latent variable. 
The KL-divergence term in Equation \ref{equation:l_vae} can also be interpreted as a regularization term.
Therefore, the VAE loss is the sum of the negative expected log
likelihood (the reconstruction error) and the regularization term \citep{larsen2016autoencoding}. 

The authors, further, propose a technique to exploit the capacity of the discriminator to differentiate between real and generated images. The capacity of a neural network is defined as an upper bound on the number of bits that can be extracted from the training data and stored in the architecture during learning \citep{baldi2019capacity}.
\cite{larsen2016autoencoding} replace the VAE reconstruction error term in Equation \ref{equation:l_vae}, for better quality images, with a reconstruction error expressed in the GAN discriminator.
For this, they introduce a Gaussian observation model for the hidden representation of the \(l\)-th layer of the discriminator \(\text{Dis}_{l}(x)\) with mean \(\text{Dis}_{l}(\Tilde{x})\) and identity covariance, 
\begin{equation*}
    p(\text{Dis}_{l}(x)|z) = \mathcal{N}(\text{Dis}_{l}(x) | \text{Dis}_{l}(\Tilde{x}), \textbf{I}),
\end{equation*}
where \(\Tilde{x} = \text{Dec}(z)\) is the output from the decoder for the data point $x$.
Subsequently, the VAE reconstruction error in Equation \ref{equation:l_vae} is replaced with the following,
\begin{equation*}
    \mathcal{L}_{\text{llike}}^{\text{Dis}_{l}} = -\mathds{E}_{z \sim Q(Z|X)}[\log  p(\text{Dis}_{l}(x)|z)].
\end{equation*}
Thus, the combined training objective for the VAE-GAN is as follows,
\begin{equation}
    \mathcal{L} = \mathcal{L}_{\text{prior}} + \mathcal{L}_{\text{llike}}^{\text{Dis}_{l}} + \mathcal{L}_{\text{GAN}}.
\end{equation}
The parameters for the decoder model \(\theta_{\text{Dec}}\) are updated weighing the decoder's reconstruction ability against the discriminator's discernment. 
The authors use a parameter, \(\gamma\), to weigh the VAE's ability to reconstruct against the discriminator. 
Therefore, the reconstruction error is,
\begin{equation*}
    \theta_{\text{Dec}} \stackrel{+}\longleftarrow  -\nabla_{\theta_{\text{Dec}}}\biggl( \gamma \mathcal{L}_{\text{llike}}^{\text{Dis}_{l}} - \mathcal{L}_{\text{GAN}}\biggr).
\end{equation*}
The training procedure for VAE-GAN is illustrated in Algorithm \ref{algorithm: 1} and Figure \ref{fig:vae_gan_train}.

\begin{figurehere}
    \centering
    \includegraphics[width=\columnwidth]{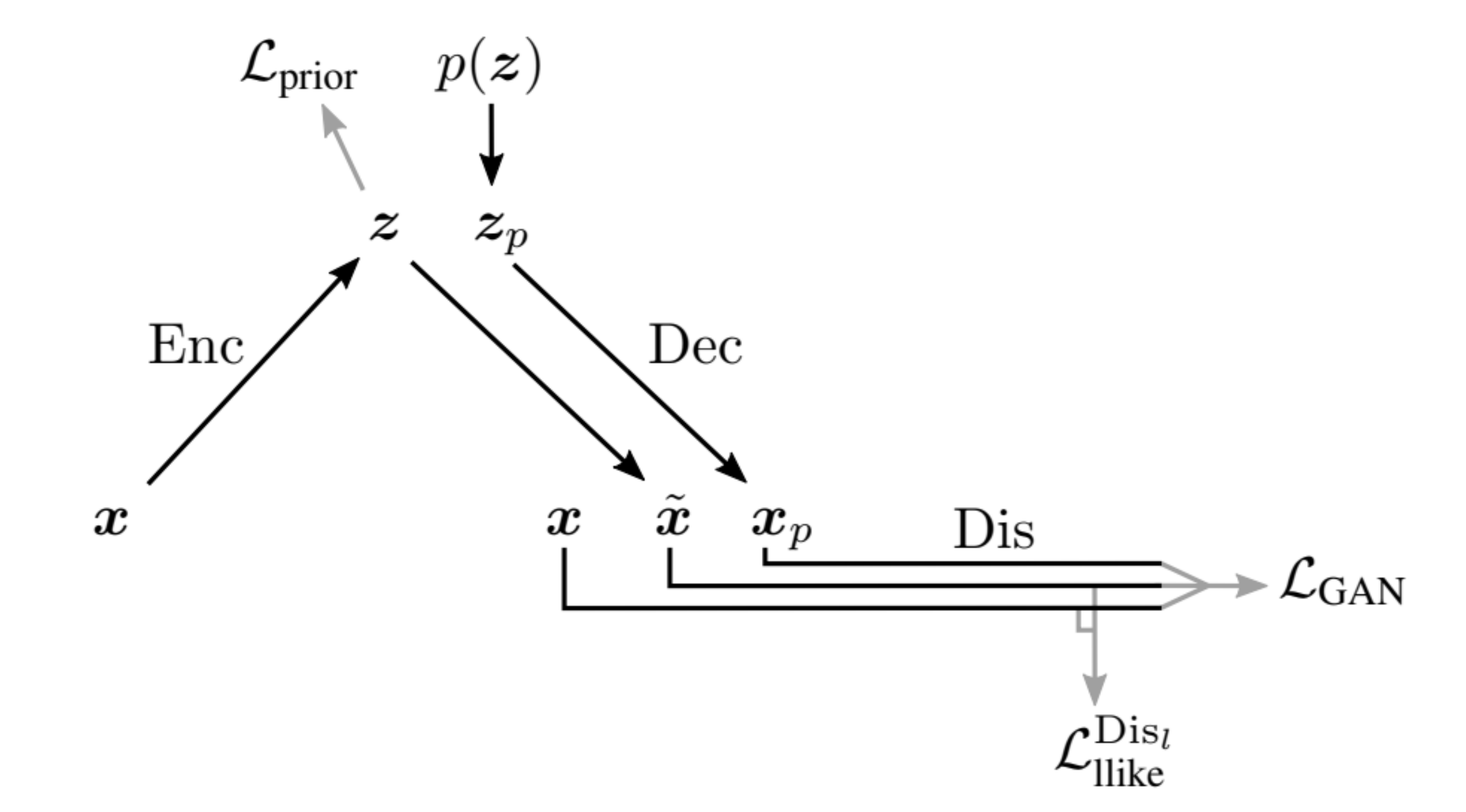}
    \caption{
    Illustration of the flow through the combined VAE-GAN framework. \cite{larsen2016autoencoding} combine a VAE with a GAN by collapsing the decoder and the generator into one. The gray lines represent the training objective.}
    \label{fig:vae_gan_train}
\end{figurehere}
\begin{algorithm}[H]
\SetAlgoLined
\DontPrintSemicolon
$\theta_{\text{Enc}}, \theta_{\text{Dec}}, \theta_{Dis} \gets$ initialize network parameters \\
\Repeat{$\textsf{\upshape convergence}$}{

    $\textbf{X} \longleftarrow$ random mini-batch from data set \\
    $\textbf{Z} \longleftarrow \text{\text{Enc}}(\textbf{X})$ \\
    $\mathcal{L}_{\text{prior}} \longleftarrow D_{\text{KL}}(Q(\textbf{Z}|\textbf{X})\;\|\;P(\textbf{Z}))$ \\
    $\Tilde{\textbf{X}} \longleftarrow \text{Dec}(\textbf{X})$ \\
    $\mathcal{L}_{\text{llike}}^{\text{Dis}_{l}} \longleftarrow -\mathds{E}_{q(\textbf{Z}|\textbf{X})}\biggl[ p(\text{Dis}_{l}(\textbf{X})|\textbf{Z})\biggr]$ \\
    $\textbf{Z}_{p} \longleftarrow$ samples from prior $\mathcal{N}(0, \textbf{I})$ \\
    $\textbf{X}_{p} \longleftarrow \text{Dec}(\textbf{Z}_{p})$ \\
    $\mathcal{L}_{\text{GAN}} \longleftarrow \log(\text{Dis}(\textbf{X}))$ \\
    $+ \log(1 - \text{Dis}(\Tilde{\textbf{X}})) +\log(1 - \text{Dis}(\textbf{X}_{p}))$ \\
    \textsc{
    \\
    Update parameters according to gradients} \\
    
    $\theta_{\text{\text{Enc}}} \stackrel{+}\longleftarrow  -\nabla_{\theta_{\text{\text{Enc}}}}\biggl( \mathcal{L}_{\text{prior}} + \mathcal{L}_{\text{llike}}^{\text{Dis}_{l}} \biggr)$ \\
    $\theta_{\text{Dec}} \stackrel{+}\longleftarrow  -\nabla_{\theta_{\text{Dec}}}\biggl( \gamma \mathcal{L}_{\text{llike}}^{\text{Dis}_{l}} - \mathcal{L}_{\text{GAN}}\biggr)$ \\
    $\theta_{\text{Dis}} \stackrel{+}\longleftarrow  -\nabla_{\theta_{\text{Dis}}}\biggl( \mathcal{L}_{\text{GAN}}\biggr)$ \\
}
\caption{Training algorithm for VAE-GAN taken from \cite{larsen2016autoencoding}}
\label{algorithm: 1}
\end{algorithm}
In recent years, the use of deep convolutional neural networks (CNNs), has resulted in state-of-the art performance for generative modeling tasks---especially in the field of computer vision \citep[e.g.][]{radford2015unsupervised, karras2019style, chen2016infogan, pidhorskyi2020adversarial}.
Such networks are computationally efficient using convolution operations to extract information from high-dimensional  data without human supervision \citep{simonyan2013deep}.
Motivated this success, \cite{larsen2016autoencoding} train CNNs along with batch-normalisation \citep{ioffe2015batch}, ReLU activations \citep{krizhevsky2012imagenet}, consecutive down- and up-sampling layers in both the encoder and discriminator.
 \cite{liu2015faceattributes} train:
\begin{itemize}
    \item[$\bullet$] a traditional VAE,
    \item[$\bullet$] a VAE with learned distance, $\text{VAE}_{\text{Dis}_l}$, where the authors first train a GAN and use the \(l\)-th layer of the discriminator network as a learned similarity measure,
    \item[$\bullet$] the proposed VAE-GAN framework,
\end{itemize}
on the CelebA data set.
As shown in Figure \ref{fig:vae_gan_results}, the visual realism of the VAE-GAN is superior to the traditional VAE. 
Additionally, the learned latent space can be used to modify high-level facial features, such as, skin tone and hair colour.

\begin{figurehere}
    \centering
    \includegraphics[width=\columnwidth]{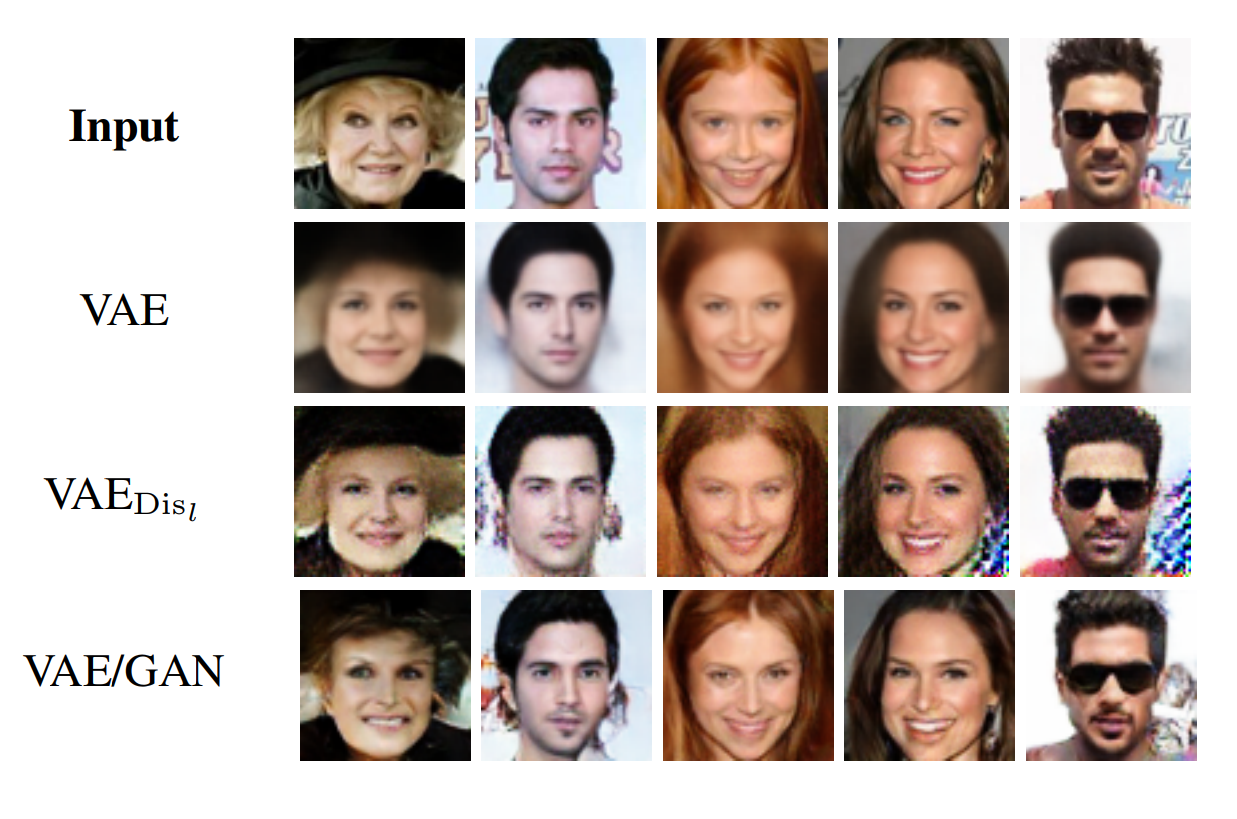}
    \caption{Reconstructions from different auto-encoders \cite{larsen2016autoencoding}.}
    \label{fig:vae_gan_results}
\end{figurehere}

The application of VI in machine learning is not limited to these frameworks.
Different variants of both VAE and VAE-GAN have been implemented and have continued to produce state of the art research in generative modelling tasks \citep[e.g.][]{zhao2019infovae, hou2017deep, vahdat2020nvae, simonovsky2018graphvae, shu2018amortized, li2020variational, purkait2020sg, zhang2018advances}. 
In addition to image generation tasks, machine learning problems like anomaly detection, time series estimation, language modelling, dimensionality reduction and unsupervised representation learning have all used VI in one form or the other \citep[e.g.][]{pol2019anomaly, li2021stacking, yang2017improved, graving2020vae}. 

\section{Discussion}
 We have introduced the concept of VI, a tool to perform approximate statistical inference. 
VI re-structures the statistical problem of estimating the posterior probability density over the latent variable, given an observed variable, into an optimization problem. 
The key idea is to select a probability density from a family of tractable densities that is closest to the actual posterior probability density. 
We have demonstrated how:
\begin{itemize}
    \item[$\bullet$] the KL-divergence can be used as a metric to measure the closeness between densities,
    \item[$\bullet$] the ELBO can be used as a criterion for model selection to better fit the observed data,
    \item[$\bullet$] VI can be used to fit a mixture of Gaussians.
\end{itemize}
Moreover, we briefly presented the scenarios where VI has been applied to modern machine learning tasks, specifically in computer vision and generative modeling, and investigated how combining deep learning and VI  enable us to perform inference on extremely complex posterior distributions.

VI is a powerful tool that allows us to approximate the actual probability density of the latent representation.
However, there are still many open avenues for statistical research.
One such avenue is to develop better approximations (achieving lower KL-divergence) to the posterior density, while maintaining efficient optimization.
For example, the mean-field family makes strong independence assumptions which aid in scalable optimization.
However, these assumptions may lead the variance of the approximated density to under-represent that of the target density \citep{blei2017variational}.
As an alternative to the mean-field method, \cite{minka2005divergence} use a fully-factorized approximation with no explicit exponential family constraint along with loopy belief propagation to achieve a lower KL-divergence.
Another possible area of research is to use \(\alpha\)-divergance measures \citep{zhang2018advances} to get a tighter fit to the ELBO.
Although research in the field of VI algorithm has grown in recent years, efforts to make VI more efficient, accurate, scalable and easier are still ongoing.


\section{Acknowledgements}
We would like to thank our colleagues Sanjana Jain, Dhruba Pujary and, Ukrit Watchareeruetai  for their constructive input and feedback during the writing of this paper.


\bibliography{references}


\cleardoublepage

\appendix
\section{Appendix}\label{section:appendix_A}
We formulate the variational approximation for the mixture of Gaussians in Equation \ref{equation:mf_gmm_init} as follows:
\begin{equation}
    q(\mu, c) = \prod_{j=1}^{K}q(\mu_{j}; m_{j}, s_{j}^2) \prod_{i=1}^{N}q(c_{i}; \phi_{i}).
    \label{equation:mf_gmm}
\end{equation}
From Equation \ref{equation:elbo_gmm__init}, we have the following definition of ELBO:
\begin{align}
    \text{ELBO}(m, s^{2}, \phi) &= \mathds{E}[\log p(x, \mu, c)] \nonumber\\
    &- \mathds{E}[\log q(\mu, c)],
    \label{equation:elbo_gmm}
\end{align}
\begin{equation*}
    \text{ELBO}_{m, s^{2}, \phi} = ELBO(m, s^{2}, \phi),
\end{equation*}
where expectations are taken under \(q\) and \(m\), \(s\) and \(\phi\) are the variational parameters.

In order to derive the optimal values of the variational parameters, we will first have to express the ELBO in Equation \ref{equation:elbo_gmm} in terms of \(m\), \(s\) and \(\phi\).
We start with simplifying the first term, \(\log p(x, \mu, c)\), on the RHS of Equation \ref{equation:elbo_gmm} as follows,
\begin{align}
    \log p(x, \mu, c) &= \log p(x | \mu, c)p(\mu, c), \nonumber \\
    &= \log p(x | \mu, c)p(\mu)p(c), \nonumber \\
    &= \log p(x | \mu, c) + \log p(\mu) + \log p(c), \nonumber \\
    &= \sum_{j} \log p(\mu_{j}) + \sum_{i}[\log p(c_{i}) \nonumber \\
    &+ \log p(x_{i} | c_{i}, \mu)],
    \label{equation:log_p_x_mu_c}
\end{align}
where \(p(c_{i}) = \frac{1}{K}\) is a constant and expanding \(p(\mu_{j})\) we have,
\begin{align}
    \log p(\mu_{j}) &= \log \bigg\{ \frac{1}{\sqrt{2\pi\sigma^2}} exp\biggl[-\frac{\mu_{j}^2}{2\sigma^2}\biggr]\bigg\} \nonumber, \\
    &\propto -\frac{\mu_{j}^2}{2\sigma^2}.
    \label{equation:log_p_mu}
\end{align}
For \(p(x_{i}|c_{i}, \mu)\), in Equation \ref{equation:log_p_x_mu_c}, we can make use of the fact that \(c_{i}\) is a one-hot vector. 
Therefore, \(\log p(x_{i}|c_{i}, \mu)\) can be expressed as:
\begin{align}
    \log p(x_{i}|c_{i}, \mu) &= \log \prod_{j} p(x_{i}| \mu_{j})^{c_{ij}}, \nonumber \\
    &= \sum_{j}c_{ij}\log p(x_{i}| \mu_{j}).
    \label{equation:log_p_x}
\end{align}
In Equation \ref{equation:true_x} we define actual density function for the real-valued data-point \(x_{i}\) as follows:
\begin{align}
    \log p(x_{i}| \mu_{j}) &= \log \bigg\{ \frac{1}{\sqrt{2\pi}} exp\biggl[ - \frac{(x_{i} - \mu_{j})^2}{2}\biggr]\bigg\}, \nonumber \\
    &\propto - \frac{(x_{i} - \mu_{j})^2}{2}.
    \label{equation:log_p_x_mu}
\end{align}
We now re-write Equation \ref{equation:log_p_x_mu_c} by combining the derivations from equations \ref{equation:log_p_mu}, \ref{equation:log_p_x} and \ref{equation:log_p_x_mu} as,
\begin{align}
    \log p(x, \mu, c) &\propto \sum_{j} -\frac{\mu_{j}^2}{2\sigma^2} \nonumber\\
    &+ \sum_{i}\sum_{j}-c_{ij}\frac{(x_{i} - \mu_{j})^2}{2}.
    \label{equation:log_p_x_mu_c_final}
\end{align}
We now factorize the variational joint probability density \(q(\mu, c)\) in Equation \ref{equation:elbo_gmm} as,
\begin{align}
    \log q(\mu, c) &= \log \prod_{j}q(\mu_{j}; m_{j}, s_{j}^2) \prod_{i}q(c_{i}, \phi_{i}), \nonumber \\
    &= \sum_{j} \log q(\mu_{j}; m_{j}, s_{j}^2) + \sum_{i} q(c_{i}, \phi_{i}).
    \label{equation:log_q_mu_c}
\end{align}
We further expand the terms on the RHS of Equation \ref{equation:log_q_mu_c} as follows,
\begin{align}
    \log q(\mu_{j}; m_{j}, s_{j}^2) &= \log \bigg\{ \frac{1}{\sqrt{2\pi s_{j}^{2}}} exp\biggl[ -\frac{(\mu_{j} - m_{j})^2}{2s_{j}^2}\biggr]\bigg\}, \nonumber \\
    &= -\frac{1}{2}\log(2\pi s_{j}^2) -\frac{(\mu_{j} - m_{j})^2}{2s_{j}^2}.
    \label{equation:log_q_mu_m_s}
\end{align}
\begin{align}
    \log q(c_{i}, \phi_{i}) &= \log \prod_{j} \phi_{ij}, \nonumber \\
    &= \sum_{j} \log \phi_{ij}.
    \label{equation:log_q_c_phi}
\end{align}
We combine the derivations for the joint variational probability density from equations \ref{equation:log_q_mu_m_s} and \ref{equation:log_q_c_phi} to re-write the Equation \ref{equation:log_q_mu_c} as,
\begin{align}
    \log q(\mu, c) &= \frac{1}{2}\sum_{j} \biggl[-\log(2\pi s_{j}^2) -\frac{(\mu_{j} - m_{j})^2}{2s_{j}^2}\biggr] \nonumber\\
    &+ \sum_{i}\sum_{j} \log \phi_{ij}.
    \label{equation:log_q_mu_c_final}
\end{align}
The final step towards deriving the ELBO in terms of the variational parameters is to factor the results from Equation \ref{equation:log_p_x_mu_c_final} and \ref{equation:log_q_mu_c_final} into Equation \ref{equation:elbo_gmm} as follows,
\begin{align*}
    \text{ELBO}_{m, s^{2}, \phi} &\propto \sum_{j}- \mathds{E}\biggl[\frac{\mu_{j}^2}{2\sigma^2}\biggr] \\
    &+\sum_{i}\sum_{j}\mathds{E}\biggl[-c_{ij}\biggr]\mathds{E}\biggl[\frac{(x_{i} - \mu_{j})^2}{2}\biggr]\\
    &-\frac{1}{2}\sum_{j} \mathds{E}\biggl[-\log(2\pi s_{j}^2) -\frac{(\mu_{j} - m_{j})^2}{s_{j}^2}\biggr] \\
    &- \sum_{i}\sum_{j} \mathds{E}\biggl[\log \phi_{ij}\biggr].
\end{align*}
The final ELBO objective in terms of the variational parameters is as follows:
\begin{align}
    \text{ELBO}_{m, s^{2}, \phi} &\propto \sum_{j} - \mathds{E}\biggl[\frac{\mu_{j}^2}{2\sigma^2}\biggr] \nonumber\\
    &+\sum_{i}\sum_{j}\mathds{E}\biggl[-\phi_{ij}\biggr]\mathds{E}\biggl[\frac{(x_{i} - \mu_{j})^2}{2}\biggr] \nonumber\\
    &+\frac{1}{2}\sum_{j}\mathds{E}\biggl[\log(s_{j}^2)\biggr] \nonumber\\
    &- \sum_{i}\sum_{j}\mathds{E}\biggl[\log \phi_{ij}\biggr],
    \label{equation:final_elbo_gmm}
\end{align}
where all expectations are taken under \(q\).

Now, to derive the optimal values of the variational parameters we take partial derivatives of the ELBO in Equation \ref{equation:final_elbo_gmm} with respect to the variational parameters and equate them to zero.

\subsection* {Deriving \(\phi_{ij}^{*}\), the optimal value of \(\phi_{ij}\):}
\begin{align*}
    \frac{\partial}{\partial \phi_{ij}}\text{ELBO}_{m, s^{2}, \phi} &\propto 
    \frac{\partial}{\partial \phi_{ij}} \Bigg\{-\phi_{ij}\mathds{E}\biggl[\frac{(x_{i} - \mu_{j})^2}{2} \biggr] \\ 
    &-\mathds{E}\biggl[\log \phi_{ij}\biggr]\Bigg\} \nonumber, \\
    &\propto \frac{\partial}{\partial \phi_{ij}} \Bigg\{-\phi_{ij}\mathds{E}\biggl[\frac{(x_{i} - \mu_{j})^2}{2} \biggr] \\ 
    &-\phi_{ij}\log \phi_{ij}\Bigg\} \nonumber, \\
    &\propto \mathds{E}\biggl[-\frac{(x_{i} - \mu_{j})^2}{2} \biggr]
    -\log \phi_{ij} - 1 \nonumber, \\
    &\propto -\frac{1}{2}(m_{j}^{2} + s_{j}^{2}) + x_{i}m_{j}
    -\log \phi_{ij}.
\end{align*}
We derive the optimal value of \(\phi_{ij}\), as follows:
\begin{align*}
    0 &= \frac{\partial}{\partial \phi_{ij}}\text{ELBO}_{m, s^{2}, \phi}, \\
    \phi_{ij}^* &\propto \exp \big[-\frac{1}{2}(m_{j}^{2}+s_{j}^{2}) + x_{i}m_{j}\big]
\end{align*}
\subsection*{Deriving \(m_{j}^{*}\), the optimal value of \(m_{j}\):}
\begin{align*}
    \frac{\partial}{\partial m_{j}}\text{ELBO}_{m, s^{2}, \phi} &\propto 
    \frac{\partial}{\partial m_{j}}
    \Bigg\{-\sum_{i}\phi_{ij}\mathds{E}\biggl[\frac{(x_{i} - \mu_{j})^2}{2} \biggr] \\
    &-\mathds{E}\biggl[\frac{\mu_{j}^2}{2\sigma^2}\biggr]\Bigg\}\nonumber \\
    &\propto \frac{\partial}{\partial m_{j}}\Bigg\{\sum_{i}\phi_{ij}
    \biggl[-\frac{1}{2}(m_{j}^{2} + s_{j}^{2}) + x_{i}m_{j}\biggr] \\ &-\frac{1}{2\sigma^{2}}(m_{j}^{2} + s_{j}^{2})\Bigg\}, \nonumber \\
    &\propto \frac{\partial}{\partial m_{j}}\Bigg\{\sum_{i}
    \biggl[-\frac{1}{2}\phi_{ij}m_{j}^{2}+ \phi_{ij}x_{i}m_{j}\biggr] \\ &-\frac{1}{2\sigma^{2}}m_{j}^{2}\Bigg\}, \nonumber \\
    &\propto \sum_{i}\biggl[\phi_{ij}m_{j}+ \phi_{ij}x_{i}\biggr] -\frac{m_{j}}{2\sigma^{2}}
\end{align*}
We derive the optimal value of \(m_{j}\), as follows:
\begin{align*}
    0 &=\frac{\partial}{\partial m_{j}}\text{ELBO}_{m, s^{2}, \phi}, \nonumber \\
    m_{j}^* &= \frac{\sum_{i} \phi_{ij}x_{i}}{\frac{1}{\sigma^{2}} + \sum_{i}\phi_{ij}}.
\end{align*}

\subsection*{Deriving \((s_{j}^{2})^*\), the optimal value of \(s_{j}^{2}\):}
\begin{align*}
    \frac{\partial}{\partial s_{j}^{2}}\text{ELBO}_{m, s^{2}, \phi} &\propto \frac{\partial}{\partial s_{j}^{2}}\Bigg\{-\sum_{i}\phi_{ij}\mathds{E}\biggl[\frac{(x_{i} - \mu_{j})^2}{2} \biggr] \\
    &-\mathds{E}\biggl[\frac{\mu_{j}^2}{2\sigma^2}\biggr] +\frac{1}{2}\mathds{E}\biggl[\log(s_{j}^2)\biggr]\Bigg\}, \nonumber \\
    &\propto \frac{\partial}{\partial s_{j}^{2}}\Bigg\{ \sum_{i}\phi{ij}\biggl[-\frac{1}{2}(m_{j}^{2} + s_{j}^{2}) + x_{i}m_{j}\biggr] \\
    &-\frac{1}{2\sigma^{2}}(m_{j}^{2} + s_{j}^{2}) + \frac{1}{2}\log(s_{j}^2) \Bigg\}, \nonumber \\
    &\propto \frac{\partial}{\partial s_{j}^{2}}\Bigg\{-\sum_{i}\frac{1}{2}\phi_{ij}s_{j}^2 - \frac{1}{2\sigma^2}s_{j}^2 + \frac{1}{2}\log s_{j}^2\Bigg\}, \nonumber \\
    &\propto -\sum_{i}\frac{1}{2}\phi_{ij} - \frac{1}{2\sigma^{2}} + \frac{1}{2s_{j}^2}.
\end{align*}
We derive the optimal value of \(s_{j}^2\), as follows:
\begin{align*}
    0 &=\frac{\partial}{\partial s_{j}^2}\text{ELBO}_{m, s^{2}, \phi}, \nonumber \\
    (s_{j}^{2})^* &= \frac{1}{\frac{1}{\sigma^{2}} + \sum_{i}\phi_{ij}}.
\end{align*}
\end{multicols}
\end{document}